# DECOMPOSITION OF FORGING DIES FOR MACHINING PLANNING

**Laurent TAPIE, Bernardin Kwamivi MAWUSSI, Bernard ANSELMETTI**

**Abstract:** *This paper will provide a method to decompose forging dies for machining planning in the case of high speed machining finishing operations. This method lies on a machining feature approach model presented in the following paper. The two main decomposition phases, called Basic Machining Features Extraction and Process Planning Generation, are presented. These two decomposition phases integrates machining resources models and expert machining knowledge to provide an outstanding process planning.*

## 1. INTRODUCTION

The recent expands of High-Speed Machining (HSM) technology and Computer-Aided Machining (CAM) software, lead forging die makers to invest in these technologies. Indeed, several works underline the productivity and part quality gains involved by HSM [1]. Yet, the difficulty inherent in machining forging dies lies in developing an outstanding machining process by using HSM. The main expert problem is to make appropriateness between die topology and machining resources (CAM software, HSM machine tool, numerical controlled unit (NCU), cutting tools, tool path…) during CAM process planning. Besides, another inherent difficulty is the die model given by forging process simulation and analysis. Indeed, in the past this analysis gave a wire frame model. According to this model machining expert defined both the die design model and the machining planning [2]. By the way of recent development in forging die process simulation, the die model is completely designed during forging simulation. Consequently, machining experts elaborate process planning starting from a 3D surface or solid Computer-Aided Design (CAD) model.

In this paper we are presenting an original approach to elaborate process planning based on the die decomposition during the finishing machining stage. This decomposition objective is to provide an outstanding process planning method taking into account productivity, part quality and resources constrains. This process planning method is divided in two stages: Basic Machining Feature Extraction (BMFE) and Process Planning Generation (PPG). In the first paper section, a forging die machining features definition model is provided in the case of three axis finishing HSM.

Then, according to this model three levels of decomposition are highlighted: geometrical and topological decomposition level, machining resources decomposition level and machining tool path decomposition level. By the way of these three decomposition levels five typical basic machining features are defined: horizontal plane feature, complex plane feature, flank contoured feature, flank followed feature and blend feature. These basic features are given by the analysis of several forging dies CAD model. First, this paper provides the machining feature model we have develop. Then, BMFE method definition is detailed in section 3. Finally, the PPG stage is provided and applied on a typical forging die.

## 2. FEATURE MODELING APPROACH

The feature modelling way is chosen in our method to associate CAD & CAM models and machining resources.

### 2.1. Forging die machining feature model

The core of this model relies on a Unified Model Language (UML) machining feature model (Fig. 1). Four principal classes are defined. The surface class is represented by the CAD die model. Typical surface used to model die part are given in [2]. The topology class represents the relationship type between machining feature. Interference topology and proximity topology are the two main topology types. This model is detailed in tab.2. The technology class gives the needed requirements for forging process. In our work this three classes are considered as data given to machining expert. The last class, called "machining process", represents the machining expert point of view. In the case of die machining this class is detailed according to the model represented in Fig. 5. Thus, the machining process represents the aggregation of three classes. The machining direction class represents the tool axis direction which can be reached according to part shape without any global interference. This class can be represented by a direction map as in [5]. In forging production the part must be extracted in one direction. Consequently, an extraction direction is associated to the part. This specificity is integrated in the technology class Fig. 1. So, in the case of forging die machining each machining feature can be machined with the same direction. Besides, three axes HSM technology is commonly used in die makers industry. Thus, the model provides a machining direction similar to the spindle axis.

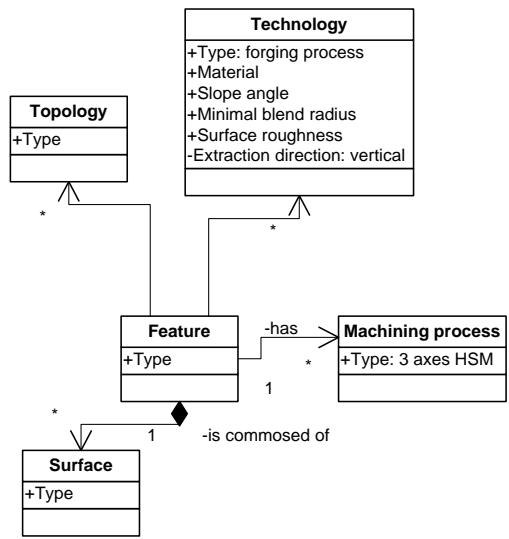

**Fig. 1.** UML machining feature model

The tool path class represents the CAM model associated to the CAD model. This class model represents the part finishing phase. Thus, this class is composed of tool trajectory during cutting phase and tool trajectory during approach or clearance motions. The cutting path is decomposed in two sub-classes: the direction path and the sweeping mode. Three common path directions are provided in CAM software for three axes finishing machining: along curves in plane containing tool axis, along curves in plane perpendicular to tool axis, along 3D curves calculated on part shape with geometric criterion (iso parametric, iso curvature…) or part quality criterion (iso scallop height). Then, a direction path is associated with a sweeping mode such as zigzag, one-way, morph in, spiral sweeping. These different data are detailed in [6].

Machining resources are also integrated in our model. The machine tool skills have an important impact in the feed rate, consequently in part quality and machining time [7]. Thus, two sub-classes are associated to the machine tool. The structure class group the machine tool data: axis stroke, maximum feed rate, maximum acceleration, jerk; spindle maximum speed, power. The NCU class groups the HSM close loop control parameter. A cinematic model of our machine tool, called "performance viewer" was implemented in Esprit CAM software [7] to model these classes. The cutting tools class model is divided in the tool cutting part and the tool body & attachment. This class is commonly used in CAM software in machining simulation to analyse local and global interference of the tools with the part shape.

*Table* 2
**Topological class model**

| | | |
|---|---|---|
| **Interference relationship** *"Junction type"* | | "lean on" or "lead into" |
| | Sub-type | Island junction<br>Cavity junction<br>Flank junction |
| | Attributes | Opened junction<br>Closed junction |
| **Interference relationship** *"Intersection type"* | | "Is in intersection with" |
| **Proximity relationship** *"Property type"* | | "belong to" |
| | Attributes | Total property<br>Partial property |
| **Proximity relationship** *"Top type"* | | "Top" |

## 3. BASIC MACHINING FEATURE EXTRACTION

### 3.1. BMFE Method

The BMFE method is composed of four main steps (Fig. 2): (1) CAM tool paths are generated in several forging die CAD models; (2) cinematic tool paths analysis is performed with CAM simulation and performance viewer model. After steps (1) and (2) machining resources decomposition level is defined by iso-resources areas. In step (3), iso-resources areas tool paths are analysed by the way of expert knowledge. This knowledge is based on some machining tests and simulation performed on our machine tool with quality and productivity criteria [6]. Then, obtained machining features are validated in step (4) and basic machining features are extracted.

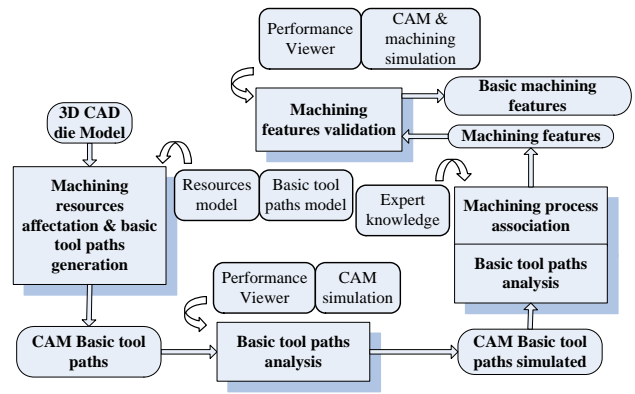

**Fig. 2.** BMFE method

### 3.2. Basic machining features models

Five typical forging die basic machining features are obtained according to BMFE method.

The "horizontal plane" (HP) feature is composed of one horizontal plane perpendicular to the tool axis. To avoid machining with a null effective cutting speed, this feature is machined with end mill tool. A surfacing machining process is associated to this feature. The sweeping mode is chosen during PPG according to topological relationship and productivity criterion.

The "complex plane" (CP) feature is composed of 1 to several surfaces. These surfaces can be represented by generated cylindrical surfaces (direction and generatrix) or composed cylindrical surfaces with surfaces primitives such as plane and revolved cylinder. The machining process associated to this feature is parallel plane machining. This machining process can be adapted during PPG according to topological relationship and productivity criterion.

The "flank contoured" (FC) feature is composed of 1 to several ruled surfaces. These surfaces can be generated or composed of primitives (cone, cylinder and plane). The inclination angle of these surfaces is closed to the slope angle given by the technology class. The machining process associated to this feature is a parallel plane contoured machining process. An adaptation of this process can be realized according to topological relationship and productivity criterion.

The "flank followed" (FF) feature is composed of 1 to several ruled surfaces. These surfaces can be generated or composed of primitives (cone, cylinder, plane). The inclination angle of these surfaces is closed to the slope angle given by the technology class. The difference between "flank contoured" is the machining process. This feature is machined along 3D surface generatrix curves.

The "blend" (B) feature is composed of blend surface. This surface assured the topological relationship between the last four features. This feature model is based on the model provided in [3]. This feature is machined along 3D curves or contoured according to the basic features in topological relation tab.3.

*Table* 3
**Blend machining feature process**

|      | CP             | FC                      | FF                      |
|------|----------------|-------------------------|-------------------------|
| SP   | No existence   | Contoured               | 3D curve                |
| CP   |                | Contoured or 3D curve   | 3D curve                |
| FC   |                |                         | Contoured or 3D curve   |

## 4. PROCESS PLANNING GENERATION (PPG)

### 4.1. PPG Method

PPG method is divided in four main steps (Fig. 3): (1) the basic features and their topological relationship are identified based on the BMFE stage results; (2) machining resources and basic machining tool path are associated based on the BMFE stage results; (3) Basic tool path are adapted according to models and experimental analyses. The basic tool paths are adapted to provide an outstanding appropriateness between machining constrains and topological constrains; (4) the process planning is generated. During step (3) some basic features are decomposed or associated, so the corresponding new features are associated with their tool paths and each feature operations are organised.

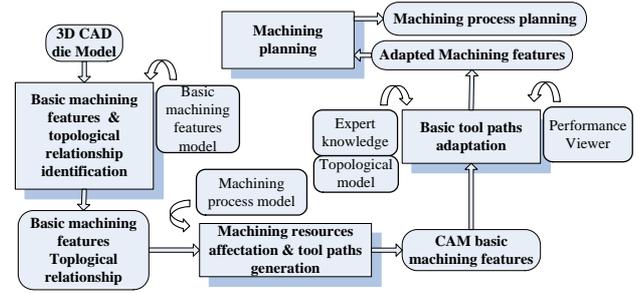

**Fig. 3.** PPG method

### 4.2. Application

Our method is applied on an industrial example. In Fig. 4, the example decomposition in basic machining feature is given. Then, topological relationship is deduced; for instance, FF1 "leans on" CP1 with an opened cavity junction and "leads into" FC2 with an opened flank junction. By the way of this decomposition each basic tool path is generated. CP1, CP2, CP3 are swept with up milling (one way) in parallel plane containing tool axis and perpendicular to the generatrix of CPi. FC1, B2 are contoured with up milling in plane parallel to CP3. FF1, FF2, B4, B5 are swept with up milling (one way) along generatix of ruled surfaces defining FF1 or FF2. FC2, FC3, FC4, B6, B7, B8, B9, B10 are contoured with Z-level up milling (one way). HP1, HP2, HP3 are surfaced with a morph out spiral and up milling (one way). B1 is swept along is generatrix curve with up milling.

Then, basic tool paths are adapted according to topological relationship and the productivity criterion given by the "performance viewer". For instance, FC1 "leans on" CP3 with a closed cavity junction. Consequently, to minimise tool retractions and approaches, CP3 is machined with a spiral from the centre to the outside limit of CP3.

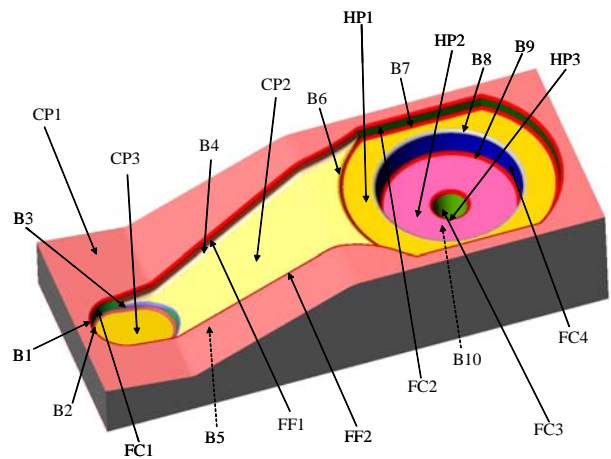

**Fig. 4.** Basic machining features identification

For CP2 basic machining process, the "performance viewer" reveals areas where the feed rate is not reached. CP3 interference topological relationship with FF1 and FF2 induces these feed rate loss. So, feed rate loss areas detected are machined with FF1 or FF2. This adaptation is validated by the "performance viewer", machine tool cinematic behaviour is better. Then, the machining process planning is totally defined (Fig. 6).

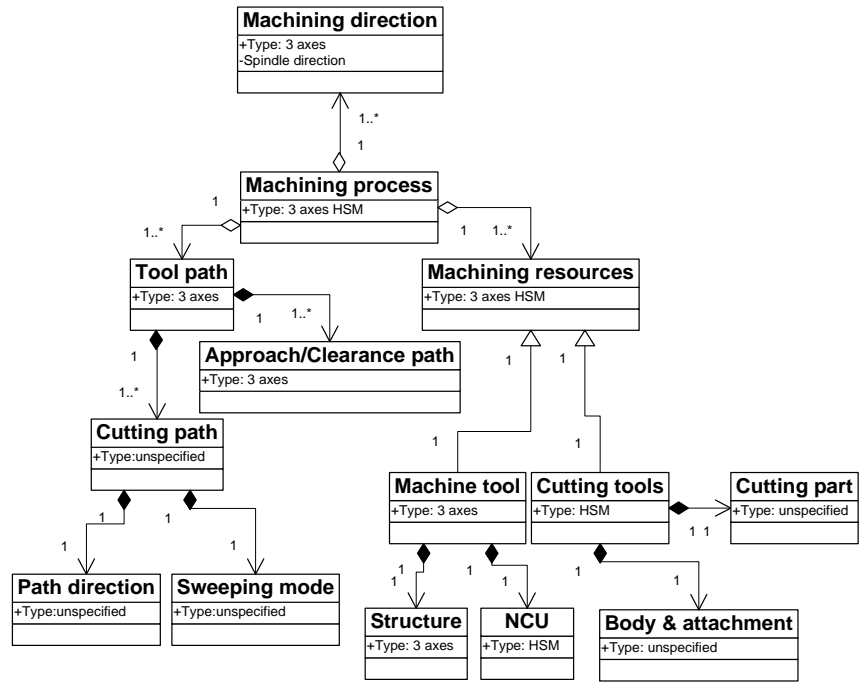

**Fig. 5.** Machining process model

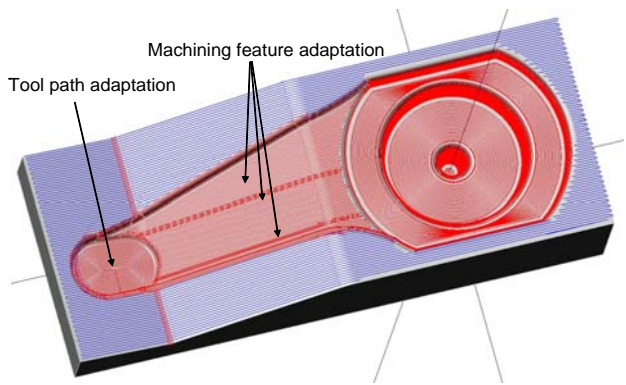

**Fig. 6.** Machining process planning.

## 5. CONCLUSION

In this paper, a method of decomposition for machining planning is provided. This method lies on a machining feature model including geometry, topology, technology of forging die and high speed machining resources. This method is divided in two stages: Basic Machining Features Extraction and Process Planning Generation.

Our future works will be focused on defining topological rules by the way of the importance of the linkage between basic machining feature topology and machining process.

Author(s):
Laurent TAPIE, PhD student, Laboratoire Universitiare de Recherche en Production Automatisée ENS-Cachan, 61, Avenue du Président Wilson, 94235 Cachan Cedex France. tapie@lurpa.ens-cachan.fr.
Dr. Bernardin Kwamivi MAWUSSI, Assistant Professor, LURPA ENS-Cachan, 61, Avenue du Président Wilson, 94235 Cachan Cedex, France. mawussi@lurpa.ens-cachan.fr.
Pr. Bernard ANSELMETTI, Professor, LURPA ENS-Cachan, 61, Avenue du Président Wilson, 94235 Cachan Cedex France. anselmetti@lurpa.ens-cachan.fr.